\documentclass[10pt,twocolumn,letterpaper]{article}

\usepackage{cvpr}              
\usepackage{balance} 
\usepackage{color}
\usepackage{float}
\usepackage{multirow}
\definecolor{cvprblue}{rgb}{0.21,0.49,0.74}
\usepackage[breaklinks,colorlinks,allcolors=cvprblue]{hyperref}
\def\confName{CVPR}
\def\confYear{2026}

\title{MMSD3.0: A Multi-Image Benchmark for Real-World Multimodal Sarcasm Detection}

\author{
Haoche Zhao$^{1,2}$ \and
Yuyao Kong$^{1,2}$ \and
Yongxiu Xu$^{1,2}$\thanks{Corresponding author} \and
Gaopeng Gou$^{1}$ \and
Hongbo Xu$^{1}$ \and
Yubin Wang$^{1,2}$ \and
Haoliang Zhang$^{1,2}$ \\
\\
$^{1}$Institute of Information Engineering, Chinese Academy of Sciences\\
$^{2}$School of Cyber Security, University of Chinese Academy of Sciences\\
\\
{\tt\small
zhaohaoche@iie.ac.cn,
xuyongxiu@iie.ac.cn,
}
}

\begin{document}
\maketitle
\begin{abstract}
Despite progress in multimodal sarcasm detection, existing datasets and methods predominantly focus on single-image scenarios, overlooking potential semantic and affective relations across multiple images. This leaves a gap in modeling cases where sarcasm is triggered by multi-image cues in real-world settings. To bridge this gap, we introduce MMSD3.0, a new benchmark composed entirely of multi-image samples curated from tweets and Amazon reviews. We further propose a \textbf{C}ross-\textbf{I}mage \textbf{R}easoning \textbf{M}odel (CIRM), integrating a Dual-Stage Bridge Module and Relevance-Guided Fusion Module to model inter-image dependencies and cross-modal correspondences. Complementarily, we establish a comprehensive suite of strong and representative baselines and conduct extensive experiments, showing that MMSD3.0 is an effective and reliable benchmark that better reflects real-world conditions. Moreover, CIRM demonstrates state-of-the-art performance across MMSD, MMSD2.0, and MMSD3.0, validating its effectiveness in both single-image and multi-image scenarios. Dataset and code are publicly available at \url{https://github.com/ZHCMOONWIND/MMSD3.0}.
\end{abstract}
\vspace{-10pt}
\section{Introduction}
\label{sec:intro}
Sarcasm is a common affective expression, yet it is hard to capture because the literal wording often contradicts the speaker’s true attitude, and it appears in diverse ironic forms.  Accurately identifying sarcasm is therefore of substantial importance for emotion detection, online services, and opinion mining~\cite{sykora2020qualitative,yin2009detection,li2024hot,xu2020reasoning,zhao2025eilmob}.

With the rapid surge of multimodal content on social media platforms, the expression of sarcasm has become increasingly diverse. In particular, sarcasm conveyed through both images and text poses significant challenges for automatic detection. This has led to the emergence of the multimodal sarcasm detection task. Cai et al.~\cite{cai2019multi} were the first to construct the widely used MMSD dataset based on Twitter, providing a benchmark for research in this area. Early approaches~\cite{cai2019multi,xu2020reasoning,liu2022towards,wen2023dip,qiao2023mutual}, built on MMSD, fused visual and textual modalities to capture cross-modal incongruity and have achieved encouraging results. However, Qin et al.~\cite{qin2023mmsd2} later identified the presence of spurious cues and problematic samples within MMSD, and subsequently introduced an improved version, MMSD2.0. This new dataset addressed these issues and further advanced the development of more reliable multimodal sarcasm detection~\cite{tang2024leveraging}.

\begin{figure}
  \centering
  \includegraphics[scale=0.43]{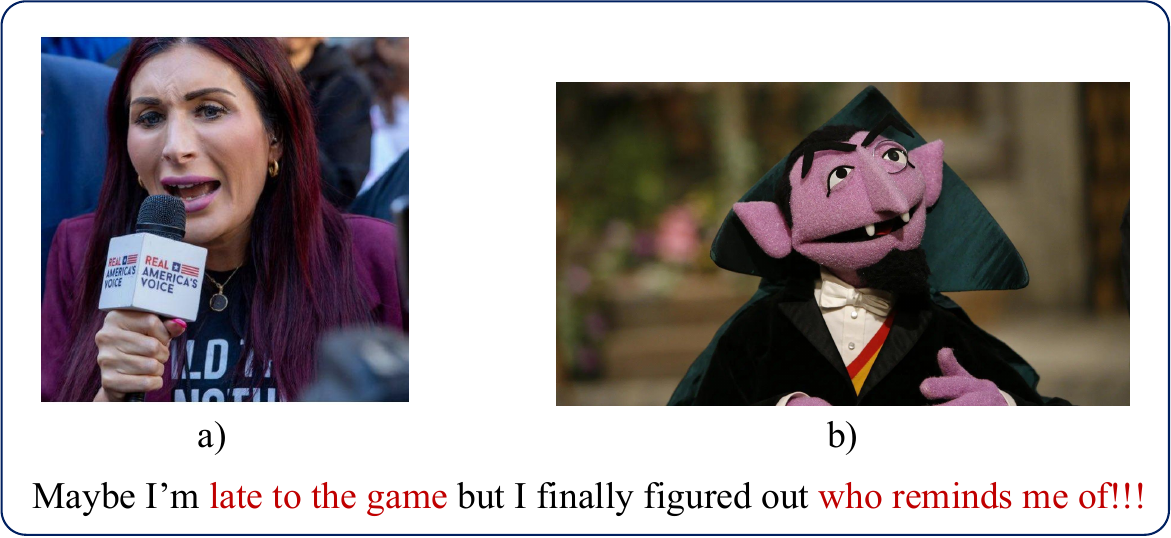}
  \caption{ An example of sarcasm created through the contrast between images. }
  \vspace{-16pt}
  \label{fig:examples}
\end{figure}

A considerable fraction of tweets contain more than one image\footnote{\url{https://thenextweb.com/news/what-analyzing-1-million-tweets-taught-us}}, which is notably high. In such multi-image scenarios, sarcasm often arises from latent semantic or affective associations among the images. As shown in Figure~\ref{fig:examples}, sarcasm can emerge from the contrast between images, for instance, the left image depicts Laura Loomer, while the right shows Count von Count. Without both images, the key contrast driving the sarcastic intent cannot be perceived. However, existing datasets~\cite{cai2019multi,qin2023mmsd2} and methods~\cite{xu2020reasoning,pan2020modeling,liu2022towards,wen2023dip,qiao2023mutual,wei2024g,chen2024cofipara,tang2024leveraging,ou2025multi,zhuang2025multi} focus solely on single-image settings and cannot capture sarcasm driven by relations across multiple images. As a result, they do not faithfully reflect the complexity of multimodal sarcasm in real-world contexts.

To advance multimodal sarcasm detection toward real-world applicability, we introduce MMSD3.0, a new dataset built from Amazon and Twitter posts. It contains over 10,000 instances, each with two to four images (the maximum allowed on Twitter). The dataset was annotated in two rounds by nine annotators to ensure high-quality labels.

At the same time, to address the challenges of multi-image sarcasm detection, we propose CIRM. CIRM features a Dual-Stage Bridge Module (DSBM) and a Relevance-Guided Fusion Module (RGFM) for efficient cross-image and cross-modal reasoning. The bridge enables information exchange before and after sequence modeling, while the fusion leverages OCR-based alignment and adaptive relevance weights to integrate visual–textual cues. Image order is further encoded through positional embeddings and masking to enhance contextual understanding.

Overall, our contributions are threefold:
\begin{itemize}
\item To the best of our knowledge, we are the first to identify the \textbf{multi-image sarcasm gap} in multimodal sarcasm detection. We propose \textbf{MMSD3.0}, a multi-image benchmark pushing the task toward real-world applicability.

\item We propose \textbf{CIRM}, a multi-image sarcasm reasoning framework that performs Dual-Stage Bridging and Relevance-Guided Fusion to enhance text–image alignment and improve multimodal understanding in complex sarcastic scenarios.

\item We conduct extensive experiments and ablations: CIRM achieves state-of-the-art results in both \textbf{single-} and \textbf{multi-image} scenarios, while prior single-image methods transfer poorly to MMSD3.0, underscoring the intrinsic difficulty of multi-image sarcasm detection.
\end{itemize}

\section{Related Work}
\label{sec:formatting}

\subsection{Multimodal Sarcasm Detection}
Early research~\cite{joshi2015harnessing,rajadesingan2015sarcasm} on sarcasm detection focused solely on the textual modality in social media. With the rise of deep learning, Zhang et al.~\cite{zhang2016tweet} proposed using a Gated Recurrent Neural Network (GRNN) for sarcasm detection, demonstrating the potential of deep learning in this task.

With multimodal content surging on social media, single‐modality sarcasm detectors are insufficient. Cai et al.~\cite{cai2019multi} established the MMSD benchmark and a hierarchical fusion model integrating text, image, and image attributes. Building on this dataset, Xu et al.~\cite{xu2020reasoning} decomposed modality‐specific representations and modeled their relations to expose cross‐modal mismatch, while Pan et al.~\cite{pan2020modeling} used self-/co-attention over BERT to capture intra-/inter‐modal incongruity. Pushing incongruity cues further, Wen et al.~\cite{wen2023dip} jointly perceived semantic and emotional discrepancies, and Qiao et al.~\cite{qiao2023mutual} coupled global–local relations via mutual-enhanced incongruity learning.

To mitigate spurious cues in MMSD, Qin et al.~\cite{qin2023mmsd2} released MMSD2.0 and proposed Multi-view CLIP to enforce image, text, and interaction views. Subsequent work refined robustness and reasoning: Zhu et al.~\cite{zhu2024tfcd} introduced training-free counterfactual debiasing (TFCD); Wei et al.~\cite{wei2024g} enhanced global semantic awareness with graph reasoning; Chen et al.~\cite{chen2024cofipara} adopted a coarse-to-fine paradigm guided by strong multimodal models. Moving to generative large-model formulations, Tang et al.~\cite{tang2024leveraging} cast detection as instruction-following with demonstration retrieval and released RedEval for OOD evaluation. Most recently, Wang et al.~\cite{wang2025rclmufn} proposed RCLMuFN, which learns relational text–image context through shallow/deep interactions and multiplex fusion (including a CLIP-view stream).

\subsection{Multimodal Large Language Models}
In multimodal research, applying powerful large language models to vision–language tasks has attracted growing attention. GPT-4o is an end-to-end “omni” model that jointly processes text and images within a single autoregressive network and demonstrates strong capabilities on VQA, captioning, chart/table understanding, and multi-image reasoning~\cite{hurst2024gpt}. LLaVA-1.5-7B adopts a data-efficient recipe by coupling a CLIP visual encoder with an MLP projection and performing visual instruction tuning on public datasets, yielding a strong open-source baseline across diverse benchmarks~\cite{liu2023improvedllava}. Qwen2.5-VL-32B emphasizes fine-grained perception via a redesigned ViT with windowed attention, native dynamic-resolution processing, and MRoPE aligned to absolute time for robust spatial–temporal grounding~\cite{Qwen2.5-VL}. Given these complementary strengths, we include them as competitive baselines in our experiments.

\begin{table}[t]
\centering
\footnotesize
\setlength{\tabcolsep}{3pt}
\caption{Dataset statistics of MMSD, MMSD2.0, and MMSD3.0 across different splits.}
\begin{tabular}{lcccccc}
\toprule
\textbf{Dataset} & \textbf{Split} & \textbf{Total} & \textbf{Sar.} & \textbf{Non-sar.} & \textbf{Sar. (\%)} & \textbf{Avg. Len / Img} \\
\midrule
\multirow{3}{*}{MMSD}      
           & Train      & 19557 & 8385 & 11172 & 42.87 & 15.65 / 1.00 \\
           & Val        & 2387 &  977 & 1410 & 40.93 & 15.68 / 1.00 \\
           & Test       & 2373 &  969 & 1404 & 40.83 & 15.82 / 1.00 \\
\midrule
\multirow{3}{*}{MMSD2.0}  
           & Train      & 19795 & 9557 & 10238 & 48.28 & 13.42 / 1.00 \\
           & Val        & 2408 & 1040 & 1368 & 43.19 & 13.64 / 1.00 \\
           & Test       & 2406 & 1034 & 1372 & 42.98 & 13.51 / 1.00 \\
\midrule
\multirow{3}{*}{MMSD3.0}      
          & Train      & 7583 & 3007 & 4576 & 39.65 & 31.81 / 2.57 \\
          & Val        & 1626 &  655 &  971 & 40.28 & 30.59 / 2.59 \\
          & Test       & 1624 &  634 &  990 & 39.04 & 33.03 / 2.61 \\
\bottomrule
\end{tabular}
\label{tab:dataset_stats}
\vspace{-12pt}
\end{table}

\vspace{-10pt}
\section{Dataset Construction}
\subsection{Data Collection}
\noindent\textbf{Out-of-domain Supplement.}
Because MMSD (Cai et al.)~\cite{cai2019multi} is sourced solely from Twitter and in-domain overfitting remains a concern~\cite{tang2024leveraging}, we supplement Twitter data\footnote{\url{https://archive.org/details/twitterstream}} with customer service–oriented reviews from Amazon\footnote{\url{https://amazon-reviews-2023.github.io/}}~\cite{hou2024bridging} to provide out-of-domain coverage.

\noindent\textbf{Data Constraints.}
MMSD labels English tweets with images that contain special hashtags (e.g., \#sarcasm) as positive and those without such hashtags as negative. This hashtag-based sampling introduces spurious cues, causing models to over-rely on text and underuse images—one key motivation for the revisions by Qin et al.~\cite{qin2023mmsd2}. Acknowledging the selection bias inherent in using special hashtags as positives, we instead build our corpus from two sources: tweets without any specific hashtags and Amazon reviews gathered without additional restrictions, thereby improving source fairness and generality.

\begin{figure}
  \centering
  \includegraphics[scale=0.17]{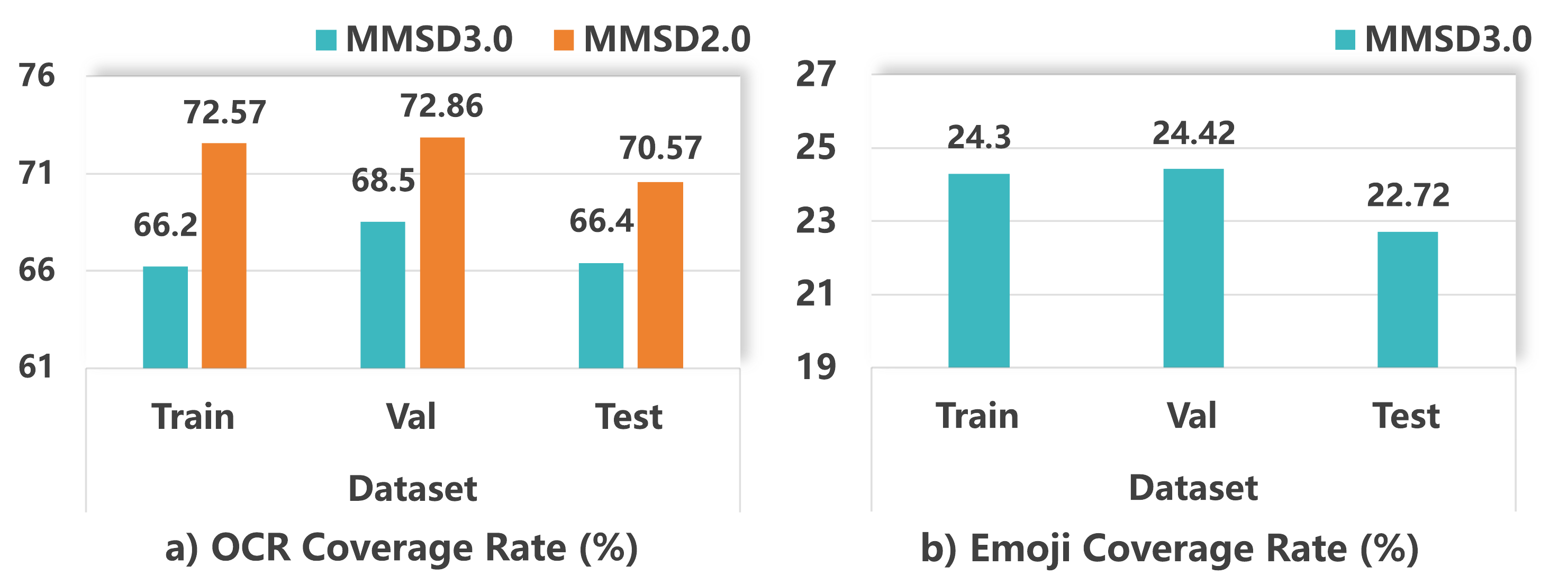}
  \caption{OCR and emoji coverage comparison of two datasets.}
  \vspace{-9pt}
  \label{fig:coverage}
\end{figure}

\noindent\textbf{Emojis.}
Emojis convey affective cues~\cite{deng2025emojis,karulkar2025emojis,ju2024emoticons}. Unlike MMSD, which replaces them with placeholders, we retain emojis to preserve sentiment signals.

\noindent\textbf{AI-generated Content and Augmentation.}
To mirror the surge of AI-generated deception on Twitter~\cite{ricker2024ai,corsi2024spread,drolsbach2025characterizing} and Amazon~\cite{yang2025classification,loke2025detecting,meng2025large}, we augment data using Qwen2.5-VL-32B~\cite{Qwen2.5-VL} as the generator and GPT-4o as the evaluator. With two platform-specific prompts (Figures~\ref{fig:tweet} and \ref{fig:amazon}) in the Appendix, we input images from 1,444 real samples to generate three sarcastic candidates per sample. As shown in Appendix Figure~\ref{fig:eval}, the evaluation prompt rates candidates on six criteria (e.g., Naturalness, Authenticity) and selects the top-scoring one, yielding 1,444 sarcasm-labeled samples.

\subsection{Human Annotation}
We recruited nine graduate students with backgrounds in multimodal research to conduct manual annotation. Given the inherently subjective nature of sarcasm~\cite{du2024docmsu,zhang2025incongruity}, each sample was assigned to two different annotators to reduce potential bias. Since understanding sarcasm in some cases may require specific background knowledge, annotators were allowed to search online for relevant context to better interpret the intended meaning. During the annotation process, we excluded any content containing violence, pornography, or other harmful or sensitive material.
To evaluate annotation consistency, we used Cohen’s Kappa~\cite{mchugh2012interrater}, and obtained a score of 0.816, indicating a high level of agreement~\cite{landis1977measurement}. Finally, we split the dataset into training, validation, and test sets following a 70:15:15 ratio.

\subsection{Data Statistics}
As shown in Figure~\ref{fig:coverage}, over 65\% of images in MMSD3.0 contain OCR-detectable text, and around 23–25\% of samples include emojis. These signals carry crucial affective and semantic cues, yet are often underutilized. By explicitly retaining them, MMSD3.0 provides a broader and more reliable pool of real-world material for sarcasm detection.

\begin{figure}
  \centering
  \includegraphics[scale=0.31]{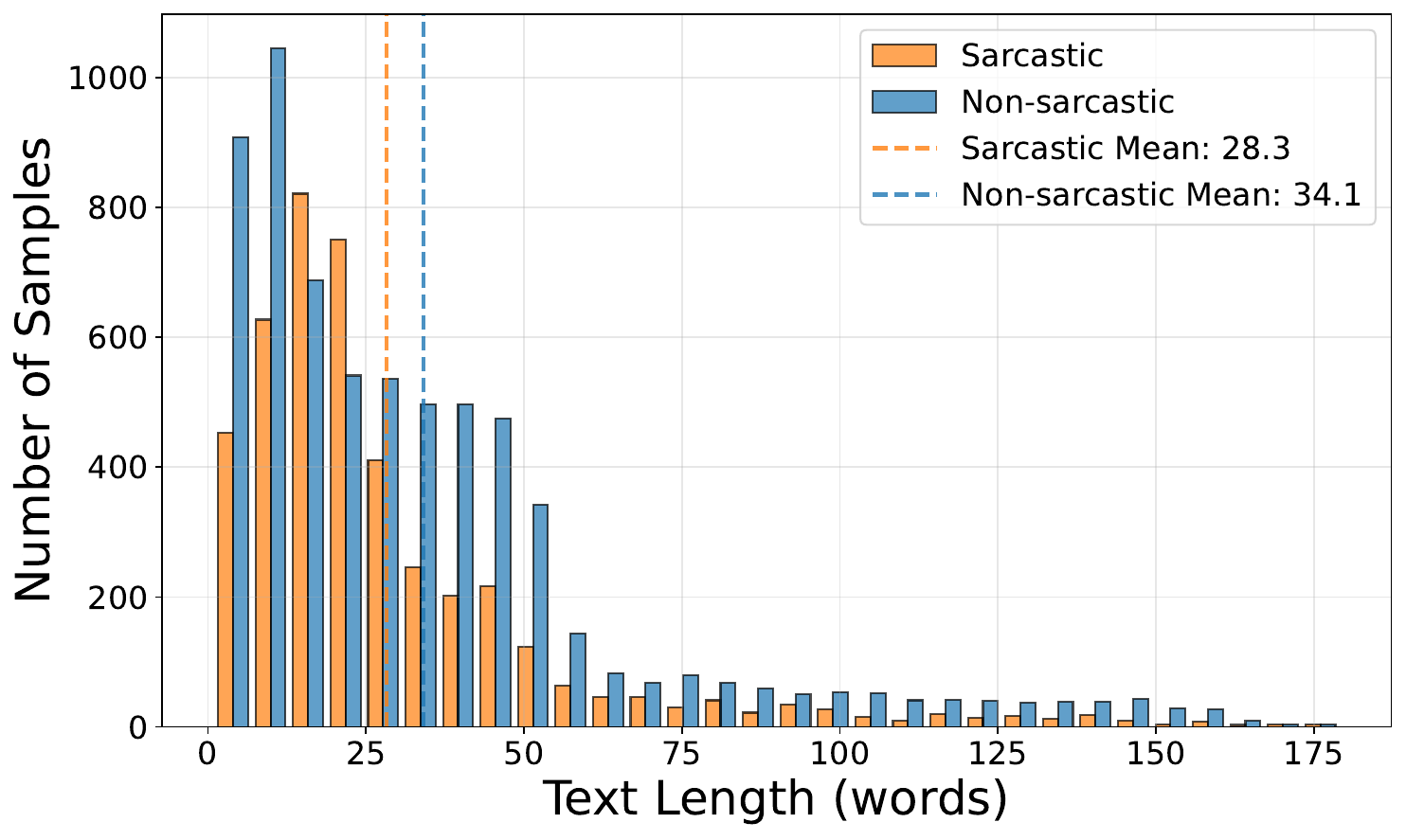}
  \caption{ Sarcastic vs Non-sarcastic Text Length Distribution. }
  \label{fig:vslen}
  \vspace{-9pt}
\end{figure}

The detailed text distribution is shown in Table~\ref{tab:dataset_stats}. Du et al.~\cite{du2024docmsu} point out that models trained on sentence-level data struggle to capture sarcasm in longer real-world texts; our dataset addresses this limitation. Compared to MMSD and MMSD2.0, which have average text lengths of 15 and 13 words, respectively, our dataset averages 31 words; the detailed text-length distribution is shown in Figure \ref{fig:vslen}. With longer contextual spans, our dataset provides richer contextual information, which can better support sarcasm detection. An additional length sensitivity evaluation using paired truncation is provided in Appendix~\ref{ex:trunc}.

Overall, MMSD3.0 more closely reflects real-world content and offers a valuable benchmark for advancing multimodal sarcasm detection in practical settings.

\section{Methodology}
\begin{figure*}
  \centering
  \includegraphics[width=\textwidth]{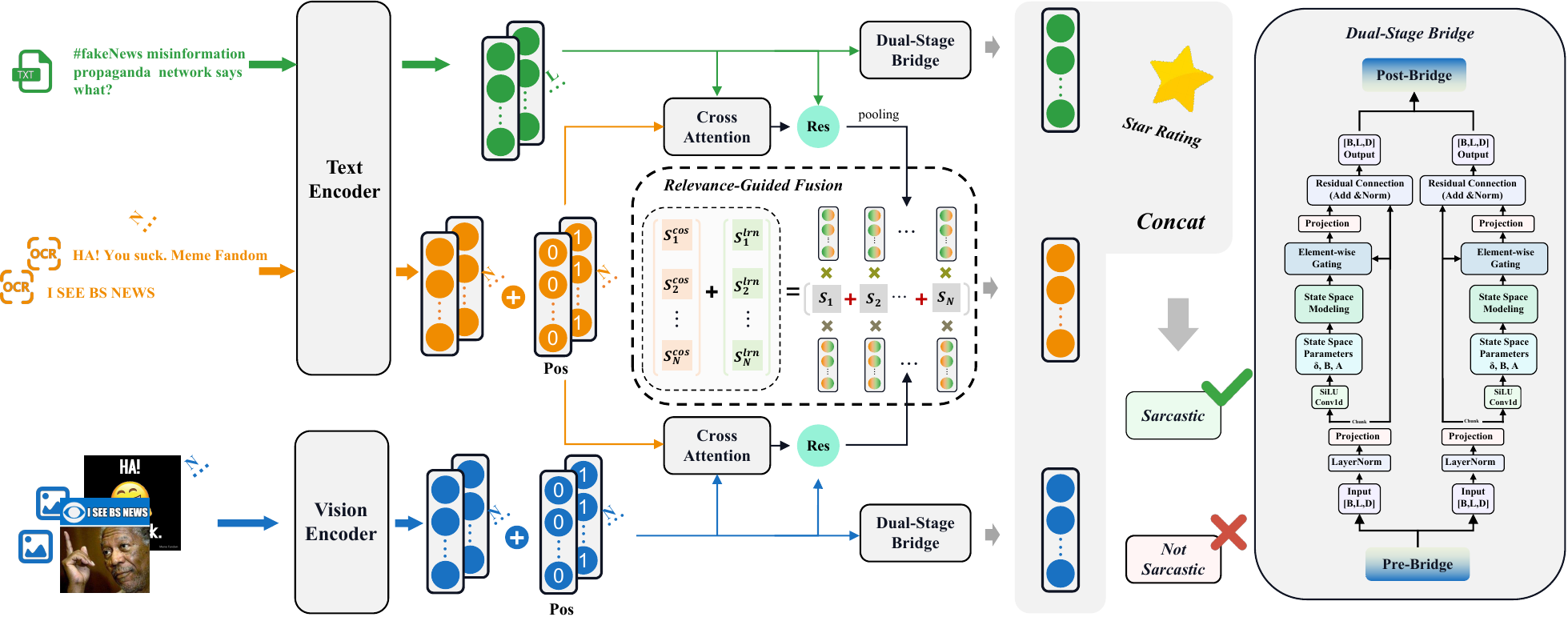}
  \caption{\centering The overall architecture of the Cross-Image Reasoning
Model}
  \label{fig:CIRM}
  \vspace{-13pt}
\end{figure*}

Figure~\ref{fig:CIRM} illustrates the overall architecture of our proposed CIRM model, which consists of five main components: Data Encoding, Positional Encoding \& Masking, Dual-Stage Bridge Module, Relevance-Guided Fusion Module, and Classification.

\subsection{Problem Formulation}
Given a text sequence \( S = (S_1, S_2, \ldots, S_L) \) and a set of associated images \( I = (I_1, I_2, \ldots, I_n) \), where \( L \) denotes the length of the text and \(n\) denotes the number of images, the task is to predict the label \( y \in \{0, 1\} \) for the sample based on the joint information from the text and the multiple images. A label of \( y = 1 \) indicates that the sample expresses sarcastic sentiment, while \( y = 0 \) indicates otherwise.

\subsection{Data Encoding}

\noindent\textbf{Image Encoding.}
We use a Vision Transformer (ViT) to encode each image. We set the maximum supported image count to $N$ = 4 (the maximum number of images allowed per post on Twitter). If a sample has fewer than four images, we pad with blank images. The encoding is:
\begin{equation}
V_{origin} = \text{Vision\_Encoder}(Pad(I))_{\texttt{[CLS]}}
\end{equation}
where \( V_{origin}=\{v^{origin}_i\}_{i=1}^{N}\) holds the CLS features of all images, and \(v^{origin}_i \in \mathbb{R}^d\) is the CLS feature of image \(I_i\).

\noindent\textbf{OCR Extraction.}
Because most images contain embedded text that conveys or shifts sentiment (e.g., memes, captions, UI overlays) and often interacts with the post caption to create cross-modal incongruity, we extract OCR using PP-OCRv5 following Cui et al.~\cite{cui2025paddleocr30technicalreport}. Let \(X=(X_1,\ldots,X_n)\), where \(X_i\) is the OCR token sequence from the \(i\)-th image and \(n\) is the number of images.

\noindent\textbf{Text \& OCR Encoding.}
Since OCR comes from images, we encode text and OCR separately instead of concatenating them~\cite{qiao2023mutual}. To leverage emojis preserved in MMSD3.0, we use RoBERTa-Emoji~\cite{barbieri2020tweeteval} as the text encoder. For input text \(S\), we take the last hidden states; for OCR \(X\), we use the \texttt{[CLS]} representations per image:
\begin{equation}
T = \text{Text\_Encoder}(S)
\end{equation}
\begin{equation}
O = \text{Text\_Encoder}(X)_{\texttt{[CLS]}}
\end{equation}
Here, \(T \in \mathbb{R}^{L \times d}\) is the representation of \(S\) with length \(L\). \(O=\{o_i\}_{i=1}^{n}\) denotes the OCR features, where \(o_i \in \mathbb{R}^d\) is the CLS vector for \(X_i\), and \(d\) is the embedding dimension.

\subsection{Positional Encoding \& Masking}

\noindent\textbf{Positional Encoding.}
Image order conveys essential contextual cues for sarcasm understanding, particularly when multiple images depict a process or contrastive narrative. 
To make the model order-aware, we add a positional embedding to each image feature according to its index in the sequence:
\begin{equation}
V = V_{\text{origin}} + \mathrm{PE}(\text{index})
\end{equation}
where \(\mathrm{PE}(\text{index}) \in \mathbb{R}^{N\times d}\) assigns each image an order-dependent embedding, 
enabling the model to capture positional and relational structure across images.

\noindent\textbf{Padding Mask.}
When fewer than $N$ images are available, blank placeholders are introduced and masked as
\begin{equation}
c = [1]^n \parallel [0]^{N-n}
\label{mask}
\end{equation}
where \( [1]^n \) and \( [0]^{N-n} \) denote all-one and all-zero vectors of length \(n\) and \(N-n\), respectively, and \( \parallel \) denotes concatenation. 
The mask \(c\) identifies valid (1) and padded (0) images, and is applied during attention and pooling to exclude non-existent inputs.

\subsection{Dual-Stage Bridge Module}

To capture cross-sequence dependencies between images and text, we propose a Dual-Stage Bridge with a Pre-Bridge and Post-Bridge, separated by a Mamba~\cite{gu2023mamba}-inspired sequential module that enhances intra- and inter-modal contextual modeling.

\noindent\textbf{Pre-Bridge.}
To enable cross-modal sequence modeling, we insert a cross-modal interaction before the Sequential Modeling Module. Let $\mathrm{MHA}(\cdot)$ be multi-head attention~\cite{vaswani2017attention} and $\mathrm{LN}(\cdot)$ layer normalization. Each modality attends to the other, and the results are fused via gated residuals:
\begin{equation}
A^{t}_{\text{pre}} = \mathrm{MHA}(T, V, V), \quad
A^{v}_{\text{pre}} = \mathrm{MHA}(V, T, T)
\end{equation}
\noindent
Here, $T$ and $V$ denote textual and visual features, respectively. Each modality attends to the other.

\begin{equation}
G^{t}_{\text{pre}} = \sigma(T W^{t}_{\text{pre}}), \quad
G^{v}_{\text{pre}} = \sigma(V W^{v}_{\text{pre}})
\end{equation}
\noindent
$W^{t}_{\text{pre}}, W^{v}_{\text{pre}} \in \mathbb{R}^{d \times d}$ are learnable projections; $\sigma$ is sigmoid. Gates $G^{t}_{\text{pre}}, G^{v}_{\text{pre}}$ control residual strengths.

\begin{equation}
T_{\text{pre}} = \mathrm{LN}(T + G^{t}_{\text{pre}} \odot A^{t}_{\text{pre}}), \quad
V_{\text{pre}} = \mathrm{LN}(V + G^{v}_{\text{pre}} \odot A^{v}_{\text{pre}})
\end{equation}
\noindent
The Hadamard product $\odot$ enables element-wise gated fusion; layer normalization stabilizes features.

\noindent\textbf{Sequential Modeling Module.}
To enhance sequence-level contextual understanding within each modality, we introduce a state-space augmented sequence block before classification. It captures local dependencies via depthwise Conv1D and models long-range interactions through selective state updates, allowing the text stream to accumulate contextual semantics and the image stream to encode coherent multi-image representations.

Given an input sequence \(H = [h_1, \dots, h_T] \in \mathbb{R}^{T \times d}\), 
we adopt a state-space inspired sequential block to strengthen intra-modal contextual modeling. 
The block first normalizes and projects features into two parallel streams:
\begin{equation}
U, Z = \mathrm{LN}(H) W_{\text{in}}, \quad U,Z \in \mathbb{R}^{T \times d}
\end{equation}
where \(U\) serves as the main information stream and \(Z\) provides dynamic gating signals.

To capture local dependencies, a depthwise convolution followed by SiLU activation is applied:
\begin{equation}
\hat{U} = \mathrm{SiLU}(\mathrm{Conv1D}(U))
\end{equation}
Then, a selective state update mechanism models long-range temporal dynamics:
\begin{equation}
S_t = f_{\theta}(S_{t-1}, \hat{u}_t)
\end{equation}
where \(f_{\theta}(\cdot)\) denotes a parameterized state transition with learnable decay and input-dependent gating, 
allowing the hidden state \(S_t\) to integrate both short-term variations and long-range context.

Finally, the updated features are gated and fused with the residual path:
\begin{equation}
H_{\text{out}} = (S \odot \mathrm{SiLU}(Z)) W_{\text{out}} + H
\end{equation}
producing the enhanced representation \(H_{\text{out}}\) that encodes hierarchical contextual dependencies within each modality.

\noindent\textbf{Post-Bridge and Final Representation.} 
After independent sequence modeling within each modality, we perform a second bridging stage to re-establish cross-modal alignment and enhance joint semantics for classification. 
The sequential block is first applied to both textual and visual streams:
\begin{equation}
T_{\text{seq}} = \mathrm{SequentialModeling}(T_{\text{pre}})
\end{equation}
\begin{equation}
V_{\text{seq}} = \mathrm{SequentialModeling}(V_{\text{pre}})
\end{equation}
Then, a gated cross-modal attention module (Post-Bridge) refines inter-modal dependencies and produces the final representations:
\begin{equation}
T_{\text{post}},\, V_{\text{post}} = 
\mathrm{CrossModalBridge}(T_{\text{seq}},\, V_{\text{seq}})
\end{equation}
This stage ensures that both modalities are contextually synchronized after their own sequence updates, yielding richer and more discriminative multimodal features for downstream classification.

\subsection{Relevance-Guided Fusion Module}
To ensure multimodal reasoning focuses on images that are semantically aligned with the textual content, 
we design a \emph{Relevance-Guided Fusion} Module. 
It first enhances both textual and visual features through OCR-guided alignment, and then computes per-image relevance to adaptively weight visual contributions, 
thereby suppressing noise from irrelevant or padded images.

\noindent\textbf{OCR-Guided Alignment.}
Given text tokens \(T \in \mathbb{R}^{L\times d}\), visual features \(V \in \mathbb{R}^{N\times d}\), and OCR embeddings \(O \in \mathbb{R}^{N\times d}\),
we inject textual grounding signals by attending to OCR cues:
\begin{equation}
T^{O}=\mathrm{Attn}(T,\,O,\,O)+T, \quad
V^{O}=\mathrm{Attn}(V,\,O,\,O)+V
\end{equation}
This process aligns both modalities with the textual context observed in the image regions.

\noindent\textbf{Relevance Estimation.}
A global textual summary \(\bar{t}_{RGFM}\) is derived by averaging \(T^{O}\). 
Each image representation \(v_i^{O}\) is then scored against \(\bar{t}\) from two perspectives:  
a cosine similarity term \(s_i^{\cos}\) and a learnable term \(s_i^{\mathrm{lrn}}\):
\begin{equation}
s_i^{\cos} = \cos(W_v v_i^{O},\, W_t \bar{t}_{RGFM}), \qquad
s_i^{\mathrm{lrn}} = \mathrm{MLP}\!\left([v_i^{O};\,\bar{t}]\right)
\end{equation}
The final relevance score mixes the two:
\begin{equation}
s_i = \alpha\, s_i^{\cos} + (1-\alpha)\, s_i^{\mathrm{lrn}}, \qquad
w_i = \mathrm{softmax}_i(s_i) \odot c_i
\end{equation}
where \(c_i \in \{0,1\}\) denotes valid (1) or padded (0) images, and \(\alpha\) balances semantic and learned relevance.

\noindent\textbf{Weighted Cross-Modal Fusion.}
The overall multimodal feature is obtained through relevance-guided aggregation:
\begin{equation}
f = \sum_{i=1}^{N} w_i\,(\bar{t} \odot v_i^{O})
\end{equation}
producing the fused representation \(f \in \mathbb{R}^{d}\) that emphasizes semantically coherent visual evidence 
while down-weighting uninformative images. 
This relevance-guided process serves as the final cross-modal integration before classification.

\begin{table*}
\centering
\caption{Comparison of performance with baseline models on the MMSD and MMSD2.0 datasets. The best results are marked in \textbf{bold}. Note that * refers to results sourced from the reference ~\cite{qin2023mmsd2}, and † refers to results cited from the original paper of the respective model.}
\small
\setlength\tabcolsep{5pt}
\begin{tabular}{cccccccccc}
\toprule
\multirow{2}{*}{\textbf{Modality}} & \multirow{2}{*}{\textbf{Method}} & \multicolumn{4}{c}{\textbf{MMSD}} & \multicolumn{4}{c}{\textbf{MMSD2.0}} \\
\cmidrule(lr){3-6} \cmidrule(lr){7-10}
& & \textbf{Acc} (\%) & \textbf{P} (\%) & \textbf{R} (\%) & \textbf{F1} (\%) & \textbf{Acc} (\%) & \textbf{P} (\%) & \textbf{R} (\%) & \textbf{F1} (\%) \\
\midrule

\multirow{3}{*}{Text-Only}
& TextCNN* ~\cite{kim2014cnn} & 80.03 & 74.29 & 76.39 & 75.32 
                                            & 71.61 & 64.62 & 75.22 & 69.52 \\
& Bi-LSTM* ~\cite{GRAVES2005602}       & 81.90 & 74.66 & 78.42 & 77.53 
                                            & 72.48 & 68.02 & 68.08 & 68.05 \\
& RoBERTa* ~\cite{liu2019roberta}            & 93.97 & 90.39 & 94.59 & 92.45 
                                            & 79.66 & 76.74 & 75.70 & 76.21 \\
\midrule

\multirow{2}{*}{Image-Only}
& ResNet* ~\cite{he2016deep}                  & 64.76 & 54.41 & 70.80 & 61.53 & 65.50 & 61.17 & 54.39 & 57.58 \\
& ViT* ~\cite{dosovitskiy2021vit} & 67.83 & 57.93 & 70.07 & 63.40 & 72.02 & 65.26 & 74.83 & 69.72 \\
\midrule

\multirow{11}{*}{Multimodal}
& HFM (ACL'19)* ~\cite{cai2019multi}                         & 83.44 & 76.57 & 84.15 & 80.18 & 70.57 & 64.84 & 69.05 & 66.88 \\
& Att-BERT (EMNLP'20)* ~\cite{pan2020modeling}              & 86.05 & 80.87 & 85.08 & 82.92 & 80.03 & 76.28 & 77.82 & 77.04 \\
& HKE (EMNLP'22)* ~\cite{liu2022towards}                   & 87.36 & 81.84 & 86.48 & 84.09 & 76.50 & 73.48 & 71.07 & 72.25 \\
& DIP (CVPR'23)† ~\cite{wen2023dip}                        & 89.59 & 88.46 & 89.13 & 89.01 & --    & --    & --    & --    \\
& Multi-view CLIP (ACL'23)† ~\cite{qin2023mmsd2}           & 88.33 & 82.66 & 88.65 & 85.55 & 85.64 & 80.33 & 88.24 & 84.10 \\
& Multi-view CLIP+TFCD (IJCAI'24)† ~\cite{zhu2024tfcd}     & 89.57 & 84.83 & 89.43 & 88.13 & 86.54 & 82.46 & 87.95 & 84.31 \\
& MoBA (MM'24)† ~\cite{xie2024moba}                        & 88.40 &  82.04 & 88.31 & 84.85 & 85.22 & 79.82 & 88.29 & 84.11 \\
& G\textsuperscript{2}SAM (AAAI'24)† ~\cite{wei2024g}      & 90.48 & 89.44 & 89.79 & 89.65 & --    & --    & --    & --    \\
& CofiPara (ACL'24)† ~\cite{chen2024cofipara}             & --    & --    & --    & --    & 85.70 & 85.96 & 85.55 & 85.89 \\
& Tang et al. (NAACL'24)† ~\cite{tang2024leveraging}      & 89.97 & 89.26 & 89.58 & 89.42 & 86.43 & 87.00 & 86.30 & 86.34 \\
& RCLMuFN (KBS'25)† ~\cite{wang2025rclmufn}               & 93.09 & 87.71 & 95.68 & 91.52 & 91.57 & 89.94 & 90.55 & 90.25 \\
\midrule

\textbf{Ours} & \textbf{CIRM} & \textbf{94.02} & \textbf{93.46} & \textbf{94.14} & \textbf{93.76} & \textbf{92.12} & \textbf{91.85} & \textbf{91.53} & \textbf{91.69} \\
\bottomrule
\end{tabular}
\label{singleimage}
\vspace{-13pt}
\end{table*}

\subsection{Classification}
After the Dual-Stage Bridge and Relevance-Guided Fusion Module, we obtain three complementary representations: 
the post-bridge text and vision streams \((T_{\text{post}}, V_{\text{post}})\) and the relevance-guided feature \(f\). 
We aggregate and integrate them in a shared fusion space for final prediction.

\noindent\textbf{Global Pooling and Fusion.}
We first derive global representations for text and vision by average pooling:
\begin{equation}
\bar{t} = \frac{1}{L}\sum_{\ell} (T_{\text{post}})_{\ell}, \quad
\bar{v} = \frac{1}{n}\sum_{i} (V_{\text{post}})_{i} \odot c_i
\end{equation}
where \(c_i \in \{0,1\}\) masks padded images. 
The pooled features \(\bar{t}\) and \(\bar{v}\) are concatenated with the relevance-guided vector \(f\) and an optional star-rating $r$ embedding:
\begin{equation}
z = \mathrm{MLP}_{\text{fuse}}\!\left([\bar{t};\,\bar{v};\,f;\,\mathrm{Emb}(r)]\right)
\end{equation}
yielding the joint multimodal representation \(z \in \mathbb{R}^{d}\).

\noindent\textbf{Classification and Loss.}
The final prediction is obtained by a linear classifier:
\begin{equation}
\hat{y} = W_{\text{cls}} z + b_{\text{cls}}
\end{equation}
and optimized with a weighted cross-entropy loss:
\begin{equation}
\mathcal{L} = \mathrm{CrossEntropy}(\hat{y}, y; w)
\end{equation}
where \(\mathbf{w}\) denotes class weights to mitigate label imbalance. 
This head unifies textual, visual, and rating cues into a compact discriminative representation for sarcasm prediction.

\section{Experiments}
\vspace{-4pt}
\noindent\textbf{Note.} For details, please refer to Appendix \S\ref{ex:no_tilling} for the no-tiling protocol, \S\ref{ex:realworld} for real-world performance, and \S\ref{ex:error} for error cases.
\vspace{-4pt}
\subsection{Datasets}

Because our data extraction differs from MMSD and MMSD2.0, we do not merge them. We first evaluated CIRM's \textbf{single-image performance} on the two public datasets, MMSD and MMSD2.0. MMSD2.0 is an improved version of MMSD, with spurious cues removed and annotations refined, using training, validation, and test splits of 80\%, 10\%, and 10\%, respectively. Subsequently, we built and evaluated multiple baselines on MMSD3.0 to assess performance in multi-image settings, with training, validation, and test splits of 70\%, 15\%, and 15\% respectively.

\subsection{Experimental Settings}

We use Qwen2.5-VL-32B~\cite{Qwen2.5-VL} as the generator and GPT-4o~\cite{hurst2024gpt} as the evaluator. During baseline implementation, we reproduce each method following its \textit{official} hyperparameters. CIRM is implemented based on the Hugging Face Transformers library and optimized using AdamW~\cite{loshchilov2019adamw}. The learning rate is set to \(2e^{-5}\) with a weight decay of \(1e^{-5}\). The batch size is 8, and the model is trained for 20 epochs. All baseline construction and model training are conducted on a single NVIDIA H100 GPU (80 GB).

\begin{table}[t]
\centering
\caption{Performance comparison with baseline models on MMSD3.0. Best results are in \textbf{bold}. Here, ``(\textit{shuffled})'' denotes results obtained by randomly permuting the image order.}

\small
\setlength\tabcolsep{1pt}
\renewcommand{\arraystretch}{1.0}

\begin{tabular}{cccccc}
\toprule
\multirow{2}{*}{\textbf{Modality}} & \multirow{2}{*}{\textbf{Method}} & \multicolumn{4}{c}{\textbf{MMSD3.0}} \\
\cmidrule(lr){3-6}
& & \textbf{Acc} (\%) & \textbf{P} (\%) & \textbf{R} (\%) & \textbf{F1} (\%) \\
\midrule

\multirow{3}{*}{Text-Only} 
& Bi-LSTM ~\cite{GRAVES2005602}       & 74.78 & 73.18 & 70.31 & 71.12 \\
& BERT ~\cite{devlin2019bert}      & 79.00 & 78.20 & 79.35 & 78.46 \\
& RoBERTa ~\cite{liu2019roberta}         & 79.99 & 79.68 & 81.15 & 79.67 \\
\midrule

\multirow{2}{*}{Image-Only}
& ResNet ~\cite{he2016deep}                 & 60.26 & 45.86 & 53.92 & 49.57 \\
& ViT ~\cite{dosovitskiy2021vit}         & 64.09 & 50.41 & 52.22 & 51.30 \\
\midrule

\multirow{4}{*}{Multimodal}
& DIP ~\cite{wen2023dip}                   & 81.83 & 78.20 & 74.13 & 76.11 \\
& Multi-view CLIP ~\cite{qin2023mmsd2}     & 81.96 & 81.84 & 81.96 & 81.96 \\
& MoBA ~\cite{xie2024moba}                 & 77.59 & 77.74 & 74.27 & 75.13 \\
& Tang et al. ~\cite{tang2024leveraging}   & 82.20 & 81.79 & 80.36 & 80.91 \\
\midrule

\multirow{3}{*}{MLLMs}
& GPT-4o ~\cite{hurst2024gpt}              & 72.62 & 71.12 & 72.42 & 71.39 \\
& LLaVA-1.5-7B ~\cite{liu2023improvedllava} & 61.12 & 60.07 & 60.82 & 59.84 \\
& Qwen2.5-VL-32B ~\cite{Qwen2.5-VL}        & 71.94 & 72.43 & 74.24 & 71.52 \\
\midrule

\multirow{2}{*}{Ours}
& CIRM (\textit{shuffled})  & 84.36 & 83.64 & 83.40 & 83.51 \\
& \textbf{CIRM} & \textbf{85.16} & \textbf{84.41} & \textbf{84.43} & \textbf{84.42} \\
\bottomrule
\end{tabular}
\label{multiimage}
\vspace{-5pt}
\end{table}

\subsection{Baselines}

\noindent\textbf{Setting.}
We evaluate \emph{CIRM} in both \emph{single-image} and \emph{multi-image} scenarios. For methods that only accept one image, we tile all $N$ images in order into a single canvas and keep the text unchanged, ensuring identical visual evidence without altering model designs.

\noindent\textbf{Single-image.}
\emph{CIRM} is evaluated against unimodal and multimodal baselines.  
\textbf{Text-only:}
TextCNN~\cite{kim2014cnn} (CNN over $n$-grams);  
Bi-LSTM~\cite{GRAVES2005602} (bidirectional RNN for long-range context);  
RoBERTa~\cite{liu2019roberta} (pretrained transformer encoder).
\textbf{Image-only:}
ResNet~\cite{he2016deep} (residual CNN);  
ViT~\cite{dosovitskiy2021vit} (Transformer over patch tokens).  
\textbf{Multimodal:}
HFM~\cite{cai2019multi} (hierarchical fusion of image, text, attributes);  
Att-BERT~\cite{pan2020modeling} (inter-modal attention on BERT);  
HKE~\cite{liu2022towards} (graph reasoning over compositional congruity);  
DIP~\cite{wen2023dip} (factual + affective cues mining);  
Multi-view CLIP~\cite{qin2023mmsd2} (diverse CLIP views for incongruity);  
Multi-view CLIP+TFCD~\cite{zhu2024tfcd} (training-free counterfactual debiasing);  
MoBA~\cite{xie2024moba} (bi-directional LoRA with MoE routing);  
G$^2$SAM~\cite{wei2024g} (graph-based global perception);  
CofiPara~\cite{chen2024cofipara} (LLM-assisted sarcasm reasoning);  
Tang et al.~\cite{tang2024leveraging} (instruction templates with demo retrieval);  
RCLMuFN~\cite{wang2025rclmufn} (relational learning with multiplex fusion).

\noindent\textbf{Multi-image.}
\emph{CIRM} and modern MLLMs natively accept multiple images; for single-image methods we use the tiling protocol above. \textbf{Text-only:}
Bi-LSTM~\cite{GRAVES2005602} (bidirectional recurrent backbone);
BERT~\cite{devlin2019bert} (masked-language Transformer baseline);
RoBERTa (TweetEval)~\cite{barbieri2020tweeteval} (robustly trained BERT variant for social text).
\textbf{Image-only:}
ResNet~\cite{he2016deep} (residual CNN on composite canvas);
ViT~\cite{dosovitskiy2021vit} (patch-token Transformer on tiled images).
\textbf{Multimodal:}
DIP~\cite{wen2023dip} (joint factual/affective fusion);
Multi-view CLIP~\cite{qin2023mmsd2} (multi-perspective CLIP alignment);
MoBA~\cite{xie2024moba} (bidirectional modality interaction for robust fusion);
Tang et al.~\cite{tang2024leveraging} (LLM prompting with retrieved demonstrations).
\textbf{MLLMs:}
GPT-4o~\cite{hurst2024gpt} (end-to-end multimodal reasoning with native multi-image input);
LLaVA-1.5-7B~\cite{liu2023improvedllava} (vision-language fine-tuning on conversational data);
Qwen2.5-VL-32B~\cite{Qwen2.5-VL} (large-scale VL model with strong perception and instruction following).
Prompts used for these MLLMs are provided in Appendix Figure~\ref{fig:MLLMs}.

\noindent\textbf{Fairness \& Reproducibility.}
We use official hyperparameters when available; trained baselines follow the same data splits and evaluation protocol as \emph{CIRM}.

\begin{table}[htbp]
\centering
\caption{Ablation experiment results on MMSD3.0. ‘w/o’ stands for ‘without’.}
\setlength\tabcolsep{5pt} 
\renewcommand{\arraystretch}{1.1} 
\begin{tabular}{ccccc}
\toprule
\textbf{Model} & \textbf{Acc (\%)} & \textbf{P (\%)} & \textbf{R (\%)} & \textbf{F1 (\%)} \\
\midrule
w/o $\mathrm{DSBM}$                & 81.90 & 81.10 & 82.32 & 81.41 \\
w/o PE                             & 83.81 & 82.91 & 83.85 & 83.25 \\
w/o OCR                            & 81.96 & 81.39 & 82.85 & 81.59 \\
w/o Emoji                          & 83.25 & 82.50 & 82.15 & 82.31 \\
w/o RGFM                            & 81.65 & 81.57 & 81.65 & 81.36 \\
w/o $\mathrm{DSBM}_{\text{pre}}$   & 82.45 & 81.51 & 82.29 & 81.81 \\
w/o $\mathrm{DSBM}_{\text{sequence}}$& 82.33 & 81.64 & 83.05 & 81.23 \\
w/o $\mathrm{DSBM}_{\text{post}}$  & 82.94 & 82.17 & 81.84 & 81.99 \\
\midrule
\textbf{CIRM} & \textbf{85.16} & \textbf{84.41} & \textbf{84.43} & \textbf{84.42} \\
\bottomrule
\end{tabular}
\label{ablation}
\vspace{-9pt}
\end{table}

\subsection{Main Results}
Based on previous experimental methodologies, we use Accuracy (Acc), Macro-Precision (P), Macro-Recall (R), and Macro-F1 score (F1) as evaluation metrics.

\noindent\textbf{Single-image Results.}  
Table~\ref{singleimage} reports single-image results on MMSD and MMSD2.0. Text-only models consistently outperform image-only ones, showing that sarcasm is mainly conveyed through language. Multimodal approaches achieve further improvements by incorporating visual cues. On MMSD2.0, \textbf{CIRM} achieves \textbf{92.12} Acc and \textbf{91.69} F1, exceeding the previous best by about 1.5 points and demonstrating strong multimodal fusion and cross-modal alignment. It also attains the highest scores on MMSD, although the dataset’s text bias~\cite{qin2023mmsd2} may slightly inflate the results. Overall, \textbf{CIRM} achieves consistent state-of-the-art performance across both datasets, confirming its effectiveness in single-image sarcasm detection.

\noindent\textbf{Multi-image Results.} 
Table~\ref{multiimage} reports results on MMSD3.0 with multiple images. Among unimodal methods, \textit{text-only} models clearly outperform \textit{image-only} ones, consistent with findings on MMSD. Bi-LSTM performs the worst, confirming the advantage of pre-trained language models. ResNet and ViT lack mechanisms for modeling cross-image relations, leading to F1 scores around 50\%. The gap between multimodal and text-only models is small, as most multimodal methods rely on single-image encoders and cannot exploit inter-image relations, making the visual signal less informative. MLLMs such as GPT-4o, LLaVA-1.5-7B, and Qwen2.5-VL-32B support multi-image input but still perform modestly, showing that sarcasm detection remains challenging even for general vision-language models. 

In contrast, CIRM achieves the best overall performance with 85.16/84.42 Acc/F1, surpassing all baselines. When the image order is randomly permuted (\textit{shuffled}), F1 drops slightly to 83.51, suggesting that \textbf{CIRM} retains strong robustness under order perturbation while still leveraging image sequence information to a meaningful extent.

\vspace{-1pt}
\subsection{Ablation Study}
\vspace{-1pt}
To verify the contribution of each component, we conduct ablation experiments under several settings: (1) \textbf{w/o DSBM}, removing the Dual-Stage Bridge Module; (2) \textbf{w/o PE}, removing positional encoding and masking; (3) \textbf{w/o OCR}, excluding OCR-based textual cues; (4) \textbf{w/o Emoji}, excluding emoji cues; (5) \textbf{w/o RGFM}, removing the Relevance-Guided Fusion Module; and (6) removing specific submodules within the Dual-Stage Bridge, including \textbf{Pre-Bridge}, \textbf{Sequential Modeling}, and \textbf{Post-Bridge}. 

Table~\ref{ablation} summarizes the results. Removing the full DSBM or the Relevance-Guided Fusion Module causes the largest decline, highlighting their crucial roles in cross-modal reasoning. Excluding positional embeddings weakens performance, showing that image order matters for multi-image understanding. Omitting the Pre-Bridge or Post-Bridge slightly reduces accuracy, confirming their auxiliary value. The absence of OCR leads to a clear drop, as many images contain text crucial for sarcasm cues. Removing emoji features causes a small decline, indicating that emojis provide complementary affective signals. Overall, these results confirm that all modules jointly enhance multimodal reasoning on MMSD3.0.

\begin{figure}
  \centering
  \vspace{-13pt}
  \includegraphics[scale=0.52]{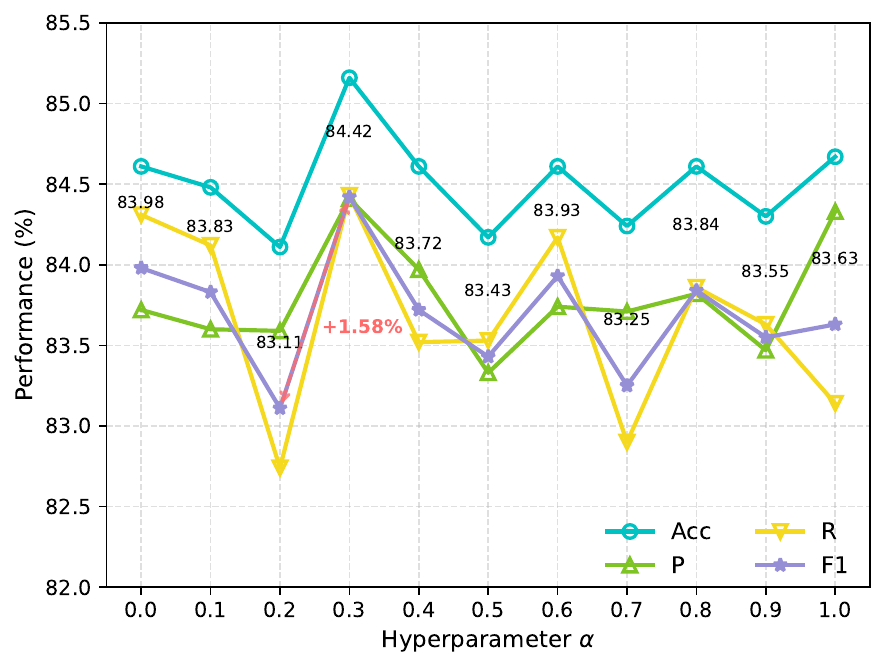}
  \vspace{-13pt}
  \caption{Effect of mixing weight \(\alpha\) on MMSD3.0, with the best F1-score (\(84.42\%\)) at \(\alpha=0.3\).}
  \label{fig:alpha}
  \vspace{-13pt}
\end{figure}

\subsection{Hyperparameter Study on Mixing Weight \texorpdfstring{$\alpha$}{alpha}}
\label{ex:alpha}
We investigate the influence of the mixing weight \(\alpha\) in the RGFM, where \(s_i=\alpha\,s_i^{\mathrm{cos}}+(1-\alpha)\,s_i^{\mathrm{lrn}}\) balances cosine alignment with the learned scorer. We sweep \(\alpha\in\{0,0.1,\ldots,1.0\}\) while keeping other settings fixed and evaluate on the MMSD3.0 test split. As shown in Figure~\ref{fig:alpha}, performance peaks at \(\alpha=0.3\), achieving the highest F1-score of \(84.42\%\). The accuracy remains stable around \(84\%\) across most values, indicating robustness to moderate changes. Both very small and large \(\alpha\) values, however, slightly reduce performance. These results suggest that a balance between cosine-based and learned relevance yields the best interaction. A small \(\alpha\) emphasizes the learned component and weakens geometric consistency, while a large one diminishes adaptive relevance signals. Considering both performance and stability, we adopt \(\alpha=0.3\) as the default in all experiments.

\begin{figure}
  \centering
  \vspace{-13pt}
  \includegraphics[scale=0.51]{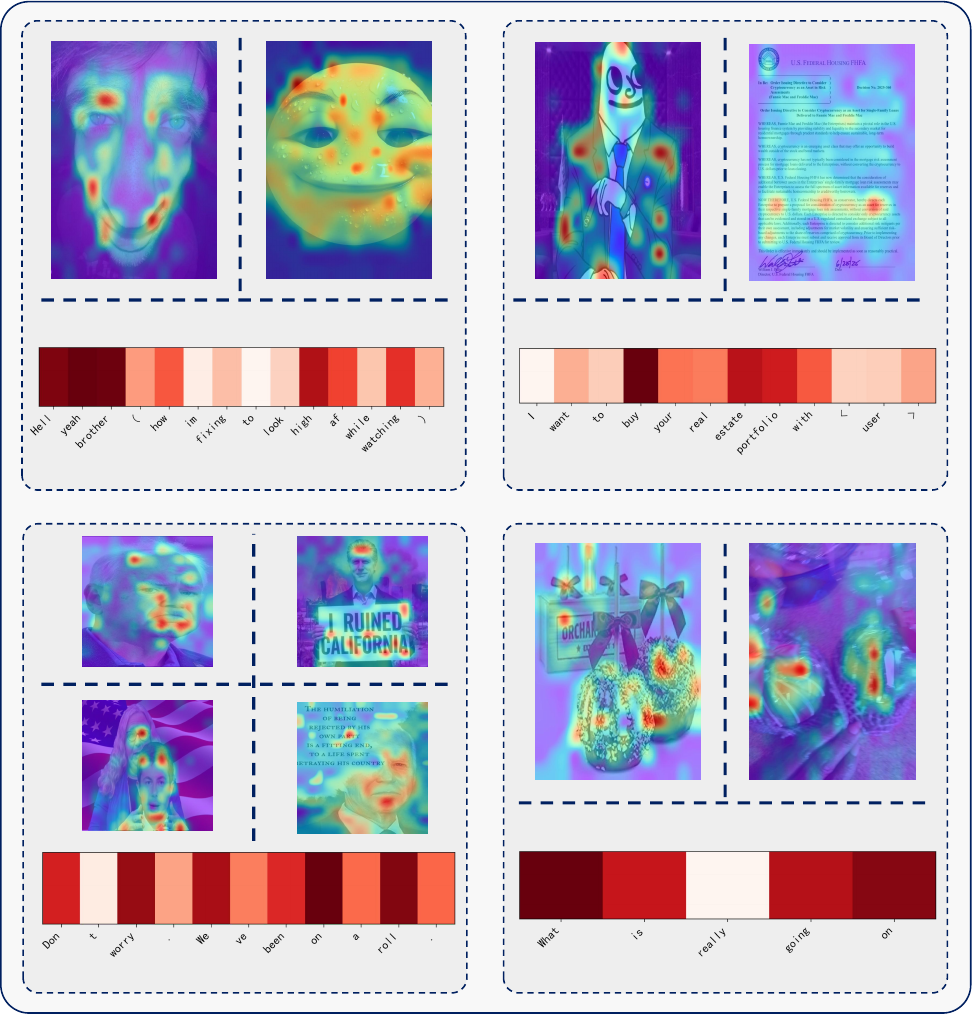}
  \vspace{-15pt}
  \caption{ Attention visualization of some examples.
Red regions in the images indicate high visual attention, green moderate, and blue low.  Text attention is shown below each image pair, with darker red denoting higher attention.}
  \label{fig:attention}
  \vspace{-8pt}
\end{figure}

\vspace{-1pt}
\subsection{Attention Visualization}
\vspace{-1pt}
Figure~\ref{fig:attention} presents qualitative attention visualizations from our CIRM model. 
The visual and textual attention maps demonstrate that the model consistently focuses on semantically meaningful regions. 
For example, in the first row, the model attends to facial regions and expressions when processing emotionally descriptive captions. 
Similarly, when the text mentions specific concepts like “real estate portfolio,” attention is directed to visually relevant areas such as documents. 
These examples suggest that CIRM captures fine-grained cross-modal alignment, effectively attending to informative content in both image and text domains, 
highlighting its capability to reason over nuanced multimodal relationships that underpin sarcasm understanding.

\vspace{-7pt}
\section*{Ethics Statement}
\vspace{-1pt}
All data used in this study are derived from publicly available datasets intended for academic research. 
We strictly follow their usage policies and ensure compliance with data privacy and content redistribution regulations. 
Personal, sensitive, or harmful information was filtered out before analysis, and no attempt was made to identify individual users. 
Annotators were informed about the task and participated voluntarily. 
To the best of our knowledge, this work does not pose any additional ethical or privacy concerns.
{
    \small
    \bibliographystyle{ieeenat_fullname}
    \bibliography{main}
}

\clearpage
\setcounter{page}{1}
\maketitlesupplementary

\begin{table}[t]
\centering
\caption{MMSD3.0 results under the modified protocol: per-image encoding with feature concatenation (no canvas tiling). Best results are in \textbf{bold}.}

\small
\setlength\tabcolsep{1pt}
\renewcommand{\arraystretch}{1.0}

\begin{tabular}{cccccc}
\toprule
\textbf{Modality} & \textbf{Method} & \multicolumn{4}{c}{\textbf{MMSD3.0}} \\
\cmidrule(lr){3-6}
& & \textbf{Acc} (\%) & \textbf{P} (\%) & \textbf{R} (\%) & \textbf{F1} (\%) \\
\midrule

\multirow{2}{*}{Image-Only}
& ResNet ~\cite{he2016deep}                          & 65.33 & 53.13 & 36.18 & 43.05 \\
& ViT ~\cite{dosovitskiy2021vit}      & 64.03 & 50.31 & 55.12 & 52.61 \\
\midrule

\multirow{4}{*}{Multimodal}
& DIP ~\cite{wen2023dip}                   & 81.49 & 81.17 & 79.95 & 80.41 \\
& Multi-view CLIP ~\cite{qin2023mmsd2}     & 82.20 & 75.86 & 79.81 & 77.79 \\
& MoBA ~\cite{xie2024moba}                 & 80.17 & 79.26 & 78.83 & 79.02 \\
& Tang et al. ~\cite{tang2024leveraging}   & 82.39 & 81.47 & 81.61 & 81.54 \\
\midrule

\multirow{1}{*}{Ours}
& \textbf{CIRM} & \textbf{85.16} & \textbf{84.41} & \textbf{84.43} & \textbf{84.42} \\
\bottomrule
\end{tabular}
\label{multiimage-modified}
\end{table}

\subsection{No-Tiling Multi-Image Experiment}
\label{ex:no_tilling}
\noindent\textbf{Setting.}
Because concatenating images can introduce unfairness, we further examine how mainstream methods perform when the framework is adapted to accept multiple images as input. Specifically, we no longer concatenate images into a single composite image. Instead, we encode each image and then concatenate the resulting feature vectors. We modify only the input format, and the model architecture remains unchanged.

\noindent\textbf{Baselines.}
\emph{CIRM} and modern MLLMs natively accept multiple images; single-image baselines are adapted by concatenating per-image features (no canvas tiling). 
\textbf{Image-only:}
ResNet~\cite{he2016deep} (residual CNN) and
ViT~\cite{dosovitskiy2021vit} (patch-token Transformer),
both using per-image encoding with feature concatenation (no canvas tiling).
\textbf{Multimodal (representative):}
DIP~\cite{wen2023dip} (joint factual/affective fusion);
Multi-view CLIP~\cite{qin2023mmsd2} (multi-perspective CLIP alignment);
MoBA~\cite{xie2024moba} (bidirectional modality interaction for robust fusion);
Tang et al.~\cite{tang2024leveraging} (LLM prompting with retrieved demonstrations).

\noindent\textbf{Fairness \& Reproducibility.}
We use official hyperparameters when available; trained baselines follow the same data splits and evaluation protocol as \emph{CIRM}.

\begin{table}[t]
\centering
\caption{Comparison of model performance on real-world and AI-generated data.}
\small
\setlength\tabcolsep{2pt}
\renewcommand{\arraystretch}{1.0}

\begin{tabular}{cccccc}
\toprule
\multirow{2}{*}{\textbf{Modality}} & \multirow{2}{*}{\textbf{Method}} & \multicolumn{2}{c}{\textbf{Real-world}} & \multicolumn{2}{c}{\textbf{AI-Gen}}\\ \cmidrule(lr){3-4} \cmidrule(lr){5-6}
& & \textbf{Acc} (\%) & \textbf{F1} (\%) & \textbf{Acc} (\%)  \\
\midrule

\multirow{4}{*}{Multimodal}
& DIP ~\cite{wen2023dip}                   & 79.59 & 75.50 & 97.98  \\
& Multi-view CLIP ~\cite{qin2023mmsd2}     & 80.01 & 75.48 & 95.96  \\
& MoBA ~\cite{xie2024moba}                 & 76.02 & 68.80 & 88.89  \\
& Tang et al. ~\cite{tang2024leveraging}   & 80.36 & 75.93 & 95.45  \\
\midrule

\multirow{3}{*}{MLLM}
& GPT-4o ~\cite{hurst2024gpt}               & 71.34 & 67.60 & 82.61  \\
& LLaVA-1.5-7B ~\cite{liu2023improvedllava} & 59.62 & 55.78 & 72.83  \\
& Qwen2.5-VL-32B ~\cite{Qwen2.5-VL}         & 68.90 & 66.65 & 95.65  \\
\midrule

\textbf{Ours} & \textbf{CIRM}  & \textbf{83.31} & \textbf{80.39} & \textbf{98.48}  \\
\bottomrule
\end{tabular}
\label{table:realworld}
\vspace{-8pt}
\end{table}

\noindent\textbf{Main Results.} 
Under the modified protocol that encodes each image separately and concatenates the resulting features, 
\textbf{CIRM} continues to achieve the highest performance on MMSD3.0 with 85.16 Acc and 84.42 F1. 
Image-only baselines such as ResNet and ViT remain weak, showing that visual cues alone are insufficient for sarcasm understanding. 
Among multimodal methods, Tang et al. and DIP retain relatively strong results, while Multi-view CLIP and MoBA show smaller gains or even reduced effectiveness, indicating that simply aggregating independent features cannot model multi-image dependencies effectively. 
Overall, these findings confirm that late feature concatenation provides limited benefit compared with architectures that perform explicit cross-image reasoning. 
\textbf{CIRM} sustains its advantage because its dual-bridging mechanism captures both inter-image relationships and cross-modal alignment, leading to a more complete multimodal understanding.

\subsection{Real-World Performance}
\label{ex:realworld}
As illustrated in Table~\ref{table:realworld}, there is a significant drop in model performance on real-world data compared to AI-generated samples, which underscores the inherent challenges posed by the greater complexity and unpredictability of real-world posts. The variation in context, language, and multimodal cues contributes to this discrepancy. Among the multimodal models, DIP and Tang et al. stand out with relatively competitive performance, whereas MoBA underperforms, and Multi-view CLIP exhibits limited adaptability to real-world conditions. On the MLLM front, while GPT-4o and Qwen2.5-VL-32B deliver moderate accuracy, LLaVA-1.5-7B lags behind in both accuracy and F1 score.

In contrast, \textbf{CIRM} consistently outperforms all baselines, achieving a robust F1 score of \(80.39\%\) on real-world data and a remarkable \(98.48\%\) on AI-generated samples. This reinforces the model’s versatility and highlights the effectiveness of its cross-image reasoning and Relevance-Guided Fusion strategies. By focusing on cross-modal consistency and latent inter-image relationships, CIRM demonstrates superior capacity to capture and interpret the subtle, multifaceted nature of sarcasm in real-world multimodal data. The results not only confirm the importance of tailored cross-image reasoning but also establish CIRM as a promising solution for multimodal sarcasm detection in complex, varied environments.

\begin{table}[t]
\centering
\normalsize
\setlength{\tabcolsep}{5pt}      
\renewcommand{\arraystretch}{1.0} 
\caption{Paired truncation on MMSD3.0 long-text ($L\!\ge\!30$, $n{=}500$) with CIRM. $\Delta$ is Full $-$ Trunc (pp).}
\begin{tabular}{lccc}
\toprule
\textbf{View} & \textbf{Acc [95\% CI]} & \textbf{P} & \textbf{R/F1} \\
\midrule
Full  & 86.40 [83.12, 89.13] & 84.67 & 83.71 / 84.16 \\
Trunc & 73.40 [69.36, 77.08] & 69.43 & 65.22 / 66.18 \\
\midrule
$\Delta$ & \textbf{+13.00} & \textbf{+15.24} & \textbf{+18.49 / +17.98} \\
\bottomrule
\end{tabular}
\label{tab:xbridge_trunc}
\end{table}

\subsection{Length Sensitivity via Paired Truncation}
\label{ex:trunc}
\noindent\textbf{Setup.}
We reuse the same \textit{CIRM} checkpoint as in the main experiments and evaluate on a randomly sampled long text subset of MMSD3.0 with $L \ge 30$ tokens ($n=500$). 
Each item is fed to the model under two input views. \emph{Full} keeps the original text. \emph{Trunc} keeps the first 15 tokens and preserves emojis and punctuation. 
Model parameters remain fixed so that the only difference arises from input length.

\noindent\textbf{Metrics.}
We report Accuracy and Macro Precision, Recall, and F1. 
Accuracy includes the Wilson 95\% confidence interval to indicate estimation uncertainty.

\noindent\textbf{Results.}
Table~\ref{tab:xbridge_trunc} shows a clear separation between the two views. 
\emph{Full} attains 86.40\% Accuracy with a 95\% confidence interval of [83.12, 89.13], and 84.67\% Macro Precision, 83.71\% Macro Recall, and 84.16\% Macro F1. 
\emph{Trunc} attains 73.40\% Accuracy with a 95\% confidence interval of [69.36, 77.08], and 69.43\% Macro Precision, 65.22\% Macro Recall, and 66.18\% Macro F1. 
The differences computed as Full minus Trunc are +13.00 percentage points on Accuracy, +15.24 on Macro Precision, +18.49 on Macro Recall, and +17.98 on Macro F1. 
The confidence intervals for Accuracy do not overlap, which indicates a statistically meaningful gap.

\noindent\textbf{Interpretation.}
Truncation removes tail spans that often contain decisive sarcasm cues such as contrastive or negated clauses, quoted punchlines, and trailing emojis or OCR referenced phrases. 
Because items and weights are identical across views, the observed degradation is attributable to the loss of long range textual evidence rather than training or sampling differences. 
The larger drops in Recall and F1 suggest that many positive instances become harder to recognize once late cues are removed, while the decline in Precision indicates increased ambiguity in shortened inputs. 
This test supports the claim that longer texts in MMSD3.0 supply essential information for robust multimodal sarcasm detection and that the proposed model effectively exploits extended context.

\begin{figure}
  \centering
  \includegraphics[scale=0.27]{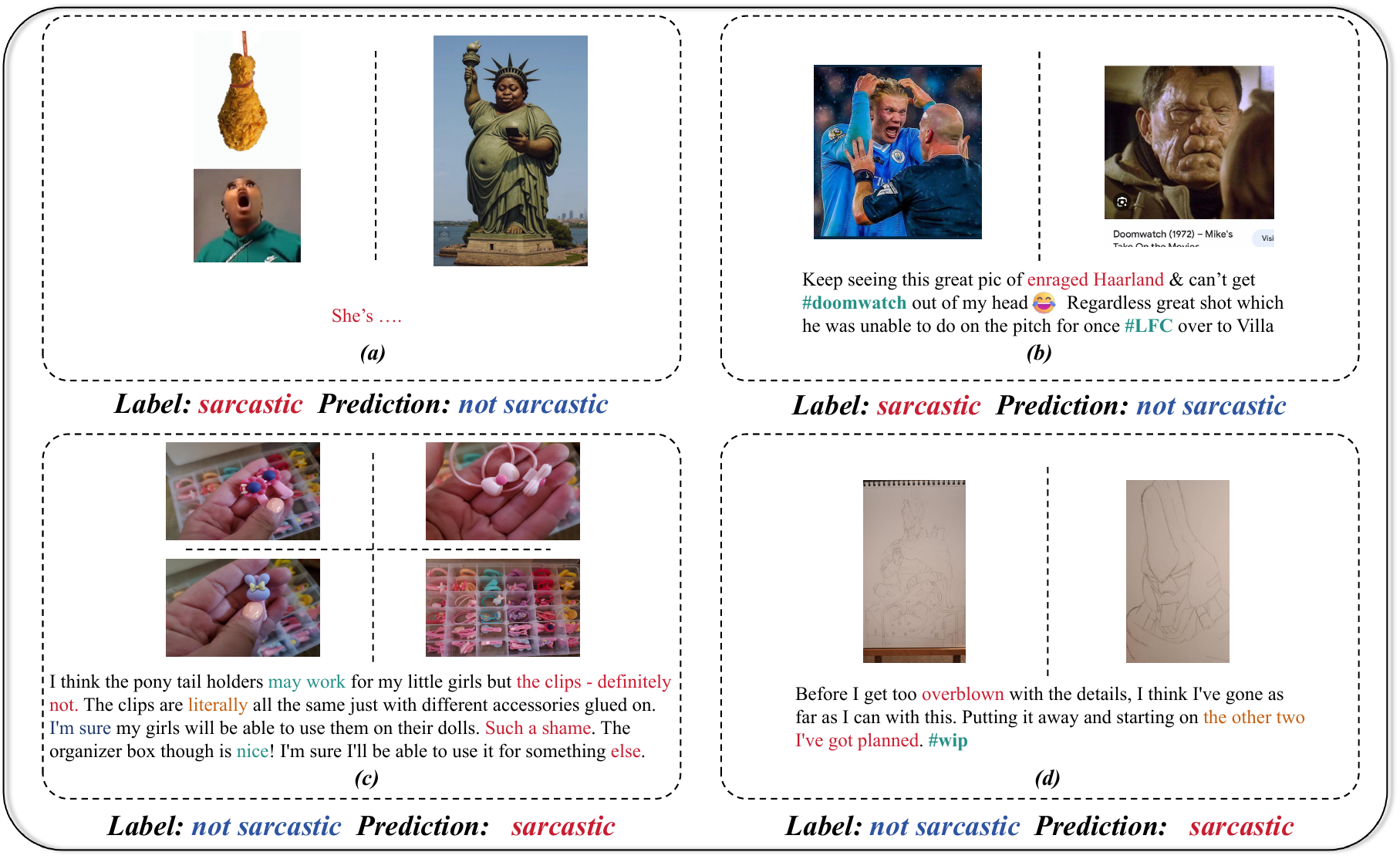}
  \caption{Error cases illustrating challenging sarcastic scenarios.}
  \label{fig:error}
  \vspace{-12pt}
\end{figure}

\subsection{Error Analysis}
\label{ex:error}
Figure~\ref{fig:error} illustrates representative failure cases revealing key challenges in multi-image sarcasm detection. In (a) and (b), the model overlooks abstract or culture-dependent incongruity. Example (a) fails mainly because the text is too short, providing no contextual signal for interpreting the visual metaphor. The sarcastic intent relies on exaggerated visual analogy across images, which the model misses without textual grounding. In (b), the model fails to recognize sarcasm connected to meme-style humor and football culture, where the reference to “doomwatch” depends on shared cultural knowledge that the model cannot infer from surface text.

In (c) and (d), subtle linguistic pragmatics—phrases like “may work,” “such a shame,” or “too overblown”—carry mixed emotional tones that confuse the sentiment classifier. These ambiguous expressions blend positive and negative cues, making it difficult for the model to correctly detect sarcasm. Overall, these cases indicate that beyond multimodal fusion, sarcasm detection requires commonsense reasoning, contextual grounding, and deeper sensitivity to pragmatic and cultural nuances.

\begin{figure}[h]
  \centering
  \includegraphics[scale=0.44]{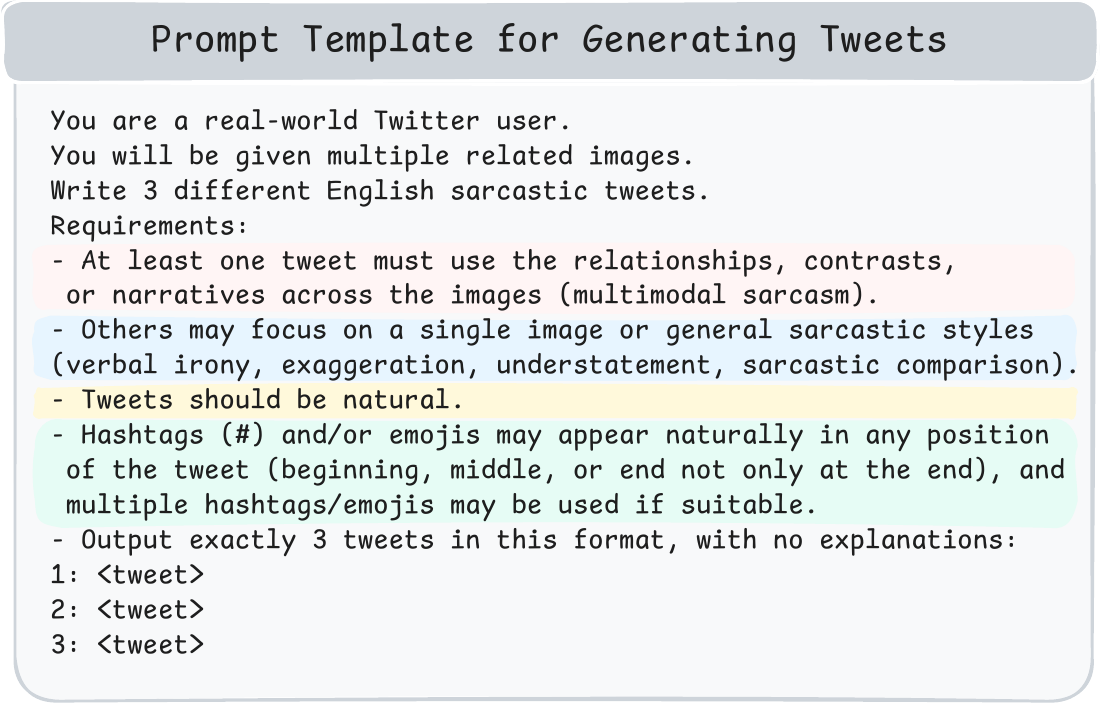}
  \caption{ Prompt Template for Generating Tweets.}
  \label{fig:tweet}
\end{figure}

\begin{figure}[h]
  \centering
  \includegraphics[scale=0.44]{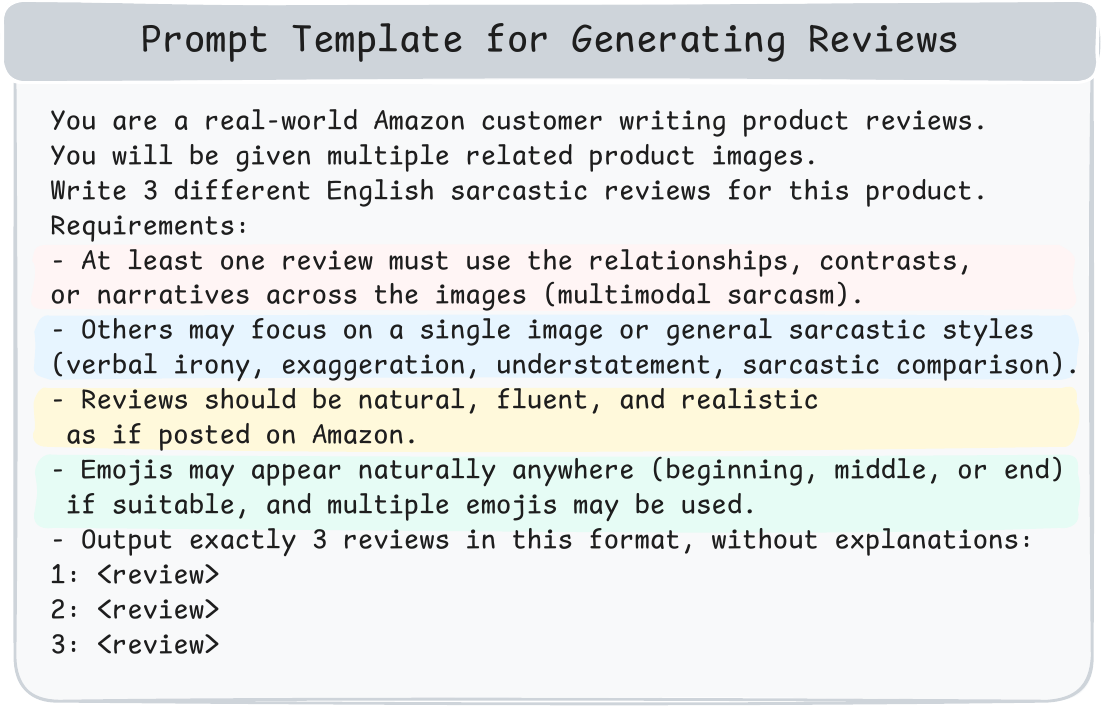}
  \vspace{-5pt}
  \caption{ Prompt Template for Generating Reviews. }
  \label{fig:amazon}
  \vspace{-5pt}
\end{figure}

\begin{figure}[h]
  \centering
  \includegraphics[scale=0.24]{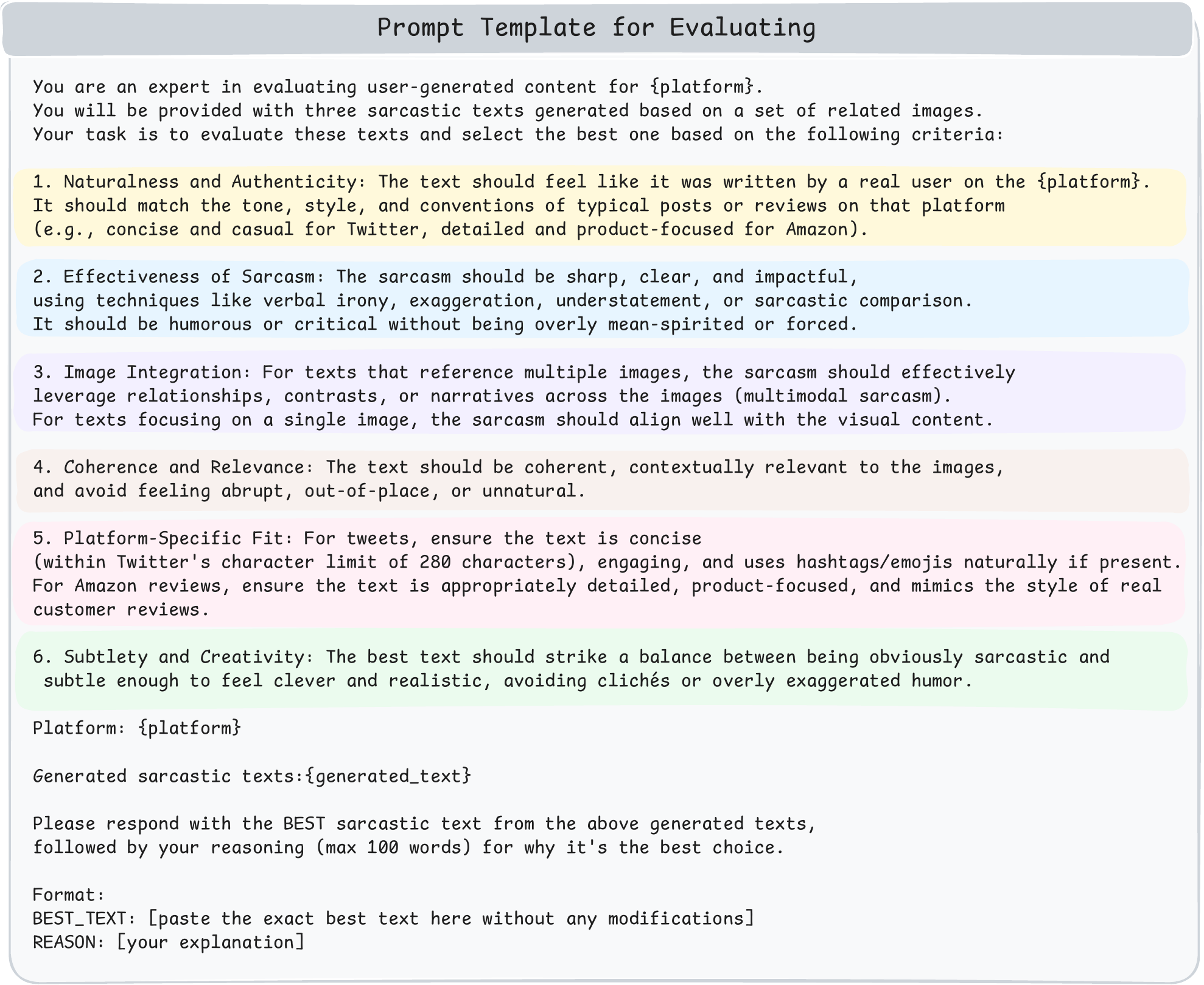}
  \vspace{-5pt}
  \caption{ Prompt Template for Evaluating}
  \label{fig:eval}
  \vspace{-5pt}
\end{figure}

\begin{figure}[h]
  \centering
  \includegraphics[scale=0.37]{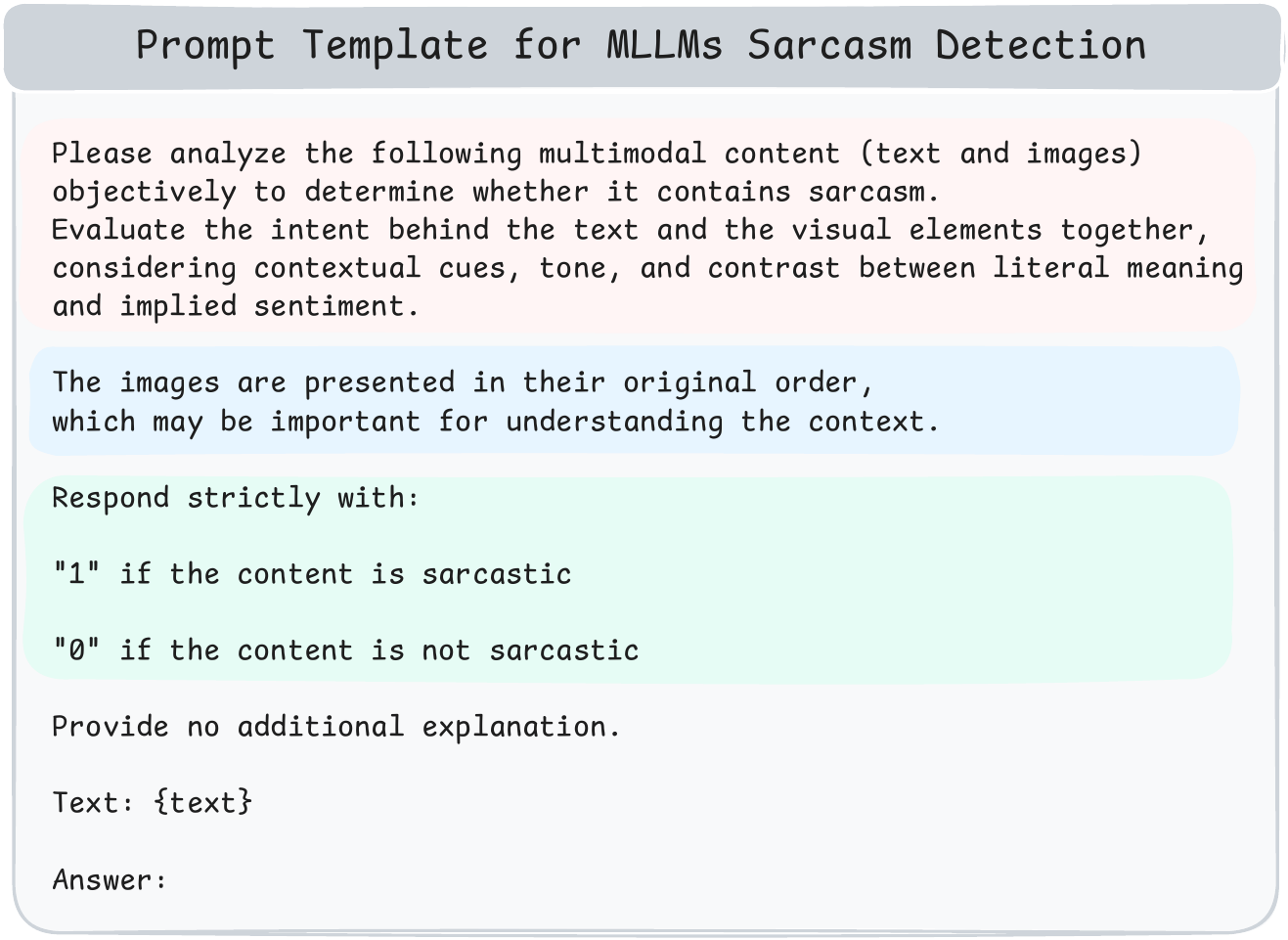}
  \vspace{-5pt}
  \caption{ Prompt Template for MLLMs Sarcasm Detection}
  \label{fig:MLLMs}
  \vspace{-5pt}
\end{figure}

\end{document}